\documentclass[11pt]{article}
\usepackage[hyperref]{ccl2022-en}
\usepackage{times}
\usepackage{url}
\usepackage{latexsym}
\usepackage{fancyhdr}
\usepackage{graphicx}
\usepackage{bm}
\usepackage{verbatimbox}
\usepackage{booktabs}
\usepackage{amsmath}
\usepackage{multirow}
\usepackage{paralist}
\usepackage{amsfonts}
\usepackage{xcolor}
\usepackage{CJKutf8}
\usepackage{enumitem}
\title{TCM-SD: A Benchmark for Probing Syndrome Differentiation via Natural Language Processing}

\author{
Mucheng Ren$^{1}$,
    \ Heyan Huang$^{1}$,\
    \ Yuxiang Zhou$^{1}$,\
    \ Qianwen Cao$^{1}$,\
    \ Yuan Bu$^{2}$,\
    and
Yang Gao$^{1}$ \\
\textsuperscript{1}School of Computer Science and Technology, Beijing Institute of Technology, Beijing, China
 \\
\textsuperscript{2}Xuzhou City Hospital of Traditional Chinese Medicine, Xuzhou, China\\
\texttt{\{renm,hhy63,yxzhou,qwcao,gyang\}@bit.edu.cn} \\
\texttt{buyuantcm@gmail.com}
}

\date{}

\begin{document}
\maketitle
\begin{abstract}
Traditional Chinese Medicine (TCM) is a natural, safe, and effective therapy that has spread and been applied worldwide. The unique TCM diagnosis and treatment system requires a comprehensive analysis of a patient's symptoms hidden in the clinical record written in free text. Prior studies have shown that this system can be informationized and intelligentized with the aid of artificial intelligence (AI) technology, such as natural language processing (NLP). However, existing datasets are not of sufficient quality nor quantity to support the further development of data-driven AI technology in TCM. 
Therefore, in this paper, we focus on the core task of the TCM diagnosis and treatment system---syndrome differentiation (SD)---and we introduce the first public large-scale benchmark for SD, called TCM-SD. Our benchmark contains 54,152 real-world clinical records covering 148 syndromes. Furthermore, we collect a large-scale unlabelled textual corpus in the field of TCM and propose a domain-specific pre-trained language model, called ZY-BERT. We conducted experiments using deep neural networks to establish a strong performance baseline, reveal various challenges in SD, and prove the potential of domain-specific pre-trained language model. Our study and analysis reveal opportunities for incorporating computer science and linguistics knowledge to explore the empirical validity of TCM theories.
\end{abstract}
\begin{CJK*}{UTF8}{gbsn}
\section{Introduction}
\label{intro}

%
%

As an essential application domain of natural language processing (NLP), medicine has received remarkable attention in recent years. Many studies have explored the integration of a variety of NLP tasks with medicine, including question answering~\cite{pampari-etal-2018-emrqa,tian-etal-2019-chimed}, machine reading comprehension~\cite{li-etal-2020-towards,yue-etal-2020-clinical}, dialogue~\cite{zeng-etal-2020-meddialog}, named entity recognition~\cite{jochim-deleris-2017-named,he-etal-2020-infusing}, and information retrieval~\cite{liu-etal-2018-know}. Meanwhile, numerous datasets in the medical domain with different task formats have also been proposed~\cite{pampari-etal-2018-emrqa,li-etal-2020-towards,tian-etal-2019-chimed}. These have greatly promoted the development of the field. Finally, breakthroughs in such tasks have led to advances in various medical-related applications, such as decision support~\cite{feng-etal-2020-explainable,panigutti2021fairlens} and International Classification of Disease (ICD) coding~\cite{cao-etal-2020-clinical,yuan2022code}.
\par
However, most existing datasets and previous studies are related to modern medicine, while traditional medicine has rarely been explored. Compared to modern medicine, traditional medicine is often faced with a lack of standards and scientific explanations, making it more challenging. Therefore, it is more urgent to adopt methods of modern science, especially NLP, to explore the principles of traditional medicine, since unstructured texts are ubiquitous in this field.

\begin{figure}[ht]
    \centering
    \includegraphics[width=0.9\linewidth,height=7cm,keepaspectratio]{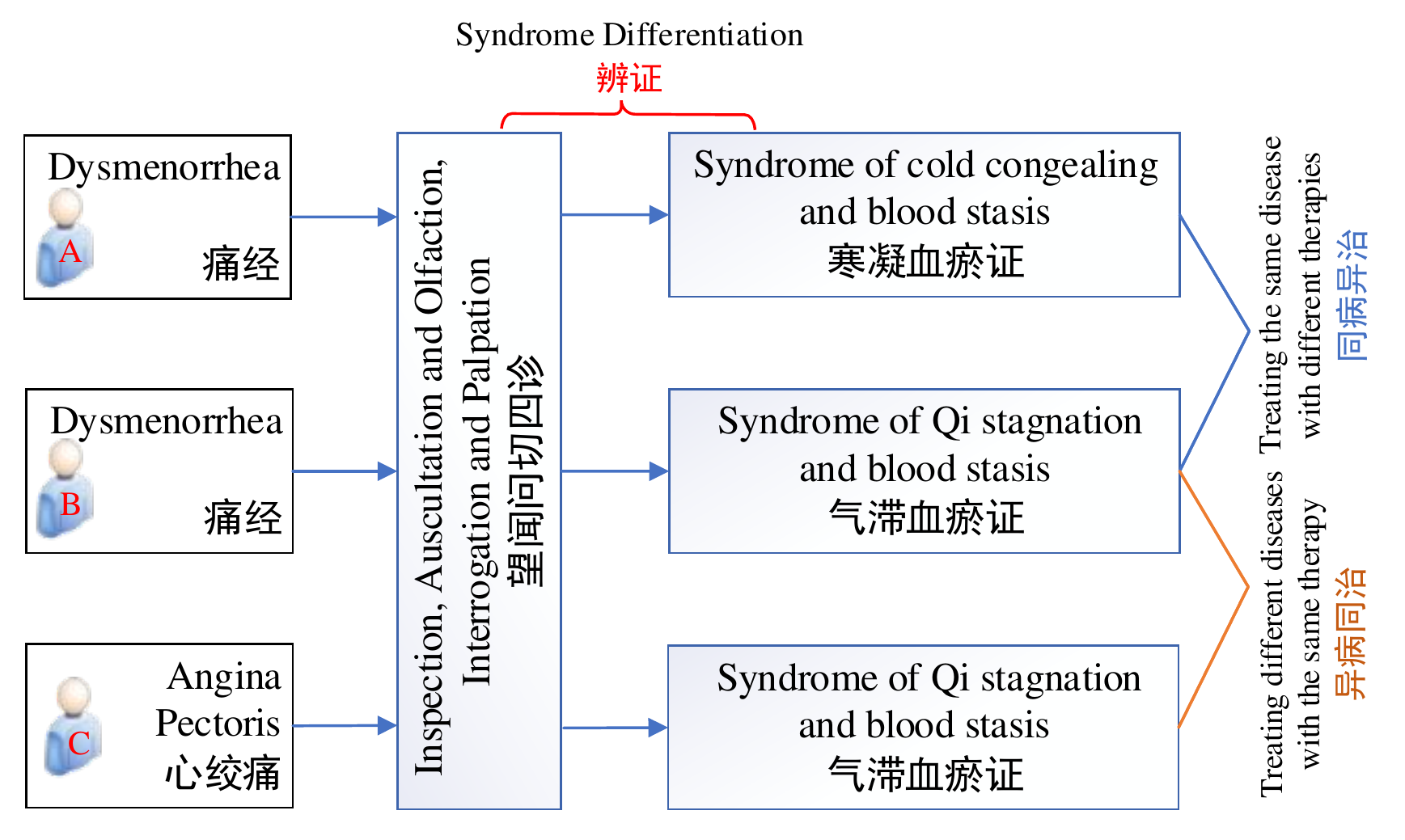}
    \caption{Concept of Traditional Chinese Medicine (TCM) syndrome differentiation.}
    \label{fig:principle}
\end{figure}
\par
TCM, as the representative of traditional medicine, is a medical system with a unique and complete theoretical basis formed by long-term medical practice under the influence and guidance of classical Chinese materialism and dialectics. Unlike modern medicine, in which medical professionals assign treatments according to disease type, TCM practitioners conduct in-depth analyses based on evidence collected from four diagnostics methods---inspection, auscultation and olfaction, interrogation, and palpation---to determine which type of \textbf{syndrome (zheng, 证)} the patient experiencing. Different treatment methods are then adopted according to the type of syndrome. Therefore, patients with the same disease may have different syndromes and thus receive different treatments, while patients with different diseases may have the same syndrome and thus undergo the same treatment. These concepts are called ``treating the same disease with different therapies (同病异治)'' and ``treating different diseases with the same therapy (异病同治),'' respectively, which are the core methods upheld by TCM. 
\par
For the example shown in Figure~\ref{fig:principle}, patients A and B have the same disease---dysmenorrhea---but one is influenced by cold while the other is driven by Qi stagnation (which is a specific concept in TCM). Thus, different therapies would be assigned. However, patient C suffered from angina pectoris but shared the same syndrome as patient B. Therefore, they would be treated with similar therapies. Thus, the \textbf{syndrome}, instead of the disease, can be regarded as the primary operating unit in the TCM medical system, which not only effectively summarizes the patients' symptoms but also determines the subsequent treatment. In this process, known as \textbf{syndrome differentiation}, \textit{the inferencing task of deciding which syndrome is associated with a patient based on clinical information}, is a vital pivot of the TCM medical system.
\par
In recent years, with the discovery of artemisinin~\cite{tu2016artemisinin} and the beneficial clinical manifestations of TCM to treat COVID-19~\cite{yang2020traditional,zhang2020becoming}, TCM has increasingly attracted attention. There have been some studies in which NLP techniques were used to explore SD tasks~\cite{zhang2019study,zhang2020artificial,wang2018cnn,liu2020end,pang2020effective}, but the development has been significantly hindered by the lack of large-scale, carefully designed, public datasets.
\par
Therefore, this paper aims to further integrate traditional medicine and artificial intelligence (AI). In particular, we focus on the core task of TCM---syndrome differentiation (SD)---to propose a high-quality, public SD benchmark that includes 54,152 samples from real-world clinical records. To our best knowledge, this is the first time that a textual benchmark has been constructed in the TCM domain. Furthermore, we crawled data from the websites to construct a TCM domain text corpus and used this to pre-train a domain-specific language model called as ZY-BERT (where ZY came from the Chinese initials of TCM). The experiments and analysis of this dataset not only explored the characteristics of SD but also verified the effectiveness of domain-specific language model. 
\par
Our contributions are summarized as follows:
\begin{enumerate}
    \item We have systematically constructed the first public large-scale SD benchmark in a format that conforms to NLP, and established the strong baselines. This can encourage researchers use NLP techniques to explore the principles of TCM that are not sufficiently explained in other fields.
    \item We proposed two novel methods, pruning and merging, which could normalize the syndrome type, improve the quality of the dataset, and also provide a reference for the construction of similar TCM datasets in the future.
    \item We proposed a domain-specific language model named as ZY-BERT pre-trained with a large-scale unlabeled TCM domain corpus, which produces the best performances so far.
\end{enumerate}

\section{Preliminaries}
To facilitate the comprehension of this paper and its motivation and significance, we will briefly define several basic concepts in TCM and analyze the differences between TCM and modern medicine.
\subsection{Characteristics of Traditional Chinese Medicine (TCM) Diagnosis}
\label{sec:tcmd}
The most apparent characteristic of TCM is that it has a unique and complete diagnostic system that differs from modern medicine. In modern medicine, with the assistance of medical instruments, the type of disease can be diagnosed according to the explicit digital indicators, such as blood pressure levels. However, TCM adopts abstract indicators, such as Yin and Yang, Exterior and Interior, Hot and Cold, and Excess and Deficiency. 
\par

\begin{figure}[ht]
    \centering
    \includegraphics[width=0.9\linewidth,height=5cm,keepaspectratio]{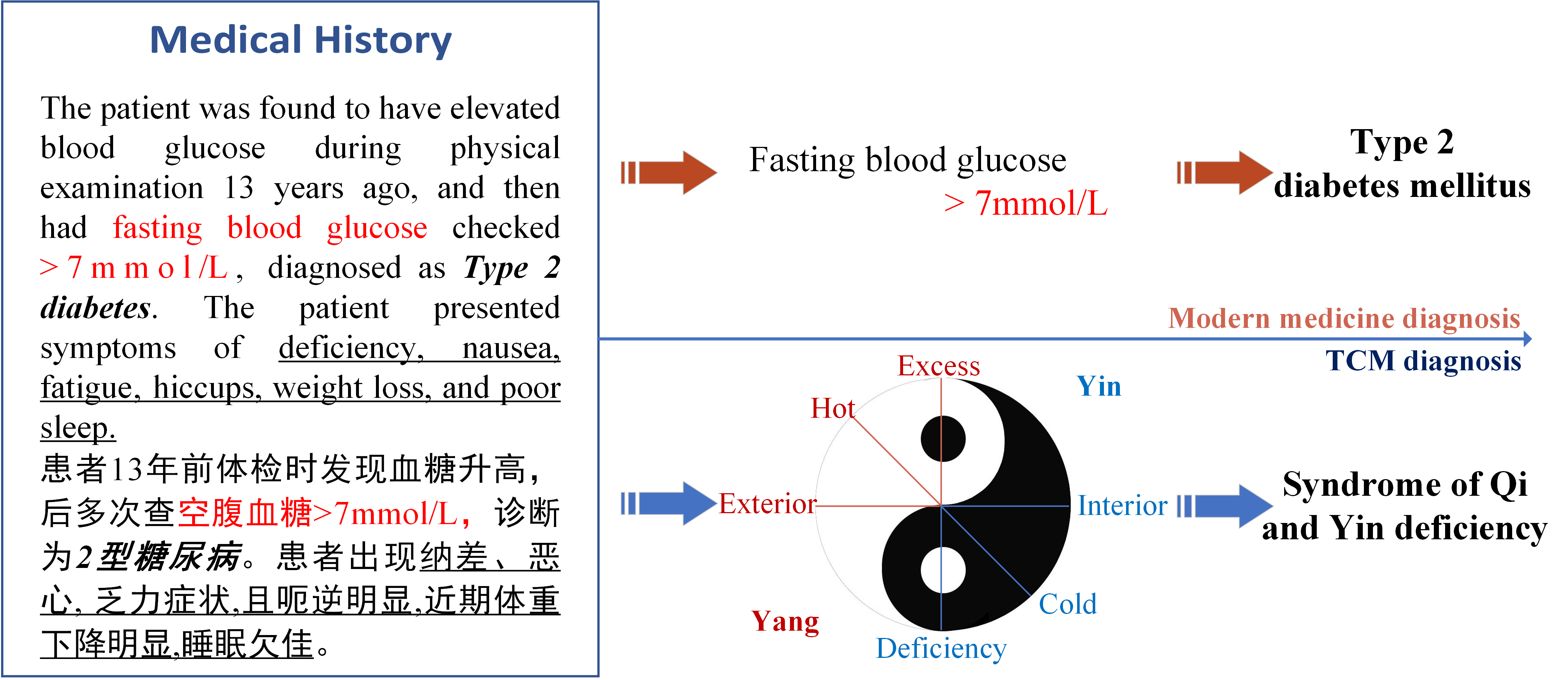}
    \caption{Different diagnostic processes of TCM and modern medicine for the same medical history. }
    \label{fig:diag}
\end{figure}
As shown in Figure~\ref{fig:diag}, given a medical history, modern medicine diagnoses the disease based on the level of fasting blood glucose, while TCM would map the various symptoms into a specific space with a unique coordinate system, analyze the latent causes, and combine them to determine a certain syndrome. Compared with the apparent numerical indicators of modern medicine, the concept of TCM is far more abstract and challenging to explain with modern medical theories. 

\par
However, TCM's difficult-to-describe nature does not mean that it has no value or rationality. In contrast, TCM has various complete and self-contained SD theories. Therefore, to explore TCM, we should not confine ourselves to the biomedical field. We may adopt NLP to explore TCM, which mainly consists of unstructured text. The linguistic characteristics may offer a scientific way to explain TCM theories. Therefore, in this paper, we present an SD dataset for further development.
\subsection{Differences between ICD coding and Syndrome Differentiation}
Automatic ICD coding is defined as assigning disease codes to Electronic Medical Records (EMR) , which is similar to TCM syndrome differentiation. Yet the two tasks are worlds apart in difficulty. Generally, the name of a patient's disease is directly recorded in EMR, and the task of the ICD coding is simply to normalize the names of these diseases in the manner of the ICD standard, without requiring a deep understanding of the context. For the example shown in Figure\ref{fig:diag}, \textbf{Type 2 diabetes} has already described in the medical history so that ICD coding can be easily completed. While the syndrome differentiation not only requires collecting scattering evidence from the context through deep understanding, but also need to execute reliable and feasible inference, which brings a huge challenge to the model.

\section{Related Works}
There are three main streams of work related to this manuscript: medical dataset, natural language processing in syndrome differentiation and domain specific pre-trained language model.
\subsection{Medical Datasets}
\begin{table*}[tb]
\resizebox{\linewidth}{!}{%
\centering
\small
\begin{tabular}{@{}lccccccc@{}}
\toprule
          & \textbf{Medical system} & \textbf{Domain} & \textbf{\# of syndromes} & \textbf{\# of samples} & \textbf{Task Type} & \textbf{Is avaliable?} & \textbf{Language} \\ \midrule
This Work & Traditional Medicine    & General         & 148                      & 54,152                 & Class.,MRC         & Yes                    & Chinese            \\ \midrule
Wang~\shortcite{wang2009feature}  & Traditional Medicine    & Liver Cirrhosis & 3                        & 406                    & Class.             & No                     & Chinese            \\
Zhang~\shortcite{zhang2019study}     & Traditional Medicine    & Stroke          & 3                        & 654                    & Class.             & No                     & Chinese            \\
Wang~\shortcite{wang2018cnn}            & Traditional Medicine    & Diabetes        & 12                       & 1,180                  & Class.             & No                     & Chinese            \\
Pang~\shortcite{pang2020effective}    & Traditional Medicine    & AIDS            & 7                        & 12,000                 & Class.             & No                     & Chinese            \\ \midrule
Johnson~\shortcite{johnson2016mimic}  & Modern Medicine         & Critical Care   & -                        & 53,423                 & -                  & Yes                    & English            \\
Stubbs~\shortcite{stubbs2015automated}      & Modern Medicine         & General         & -                        & 1,304                  & De-ID.             & Yes                    & English            \\
Dougan~\shortcite{dougan2014ncbi}     & Modern Medicine         & General         & -                        & 6,892                  & DNR                & Yes                    & English            \\
Abacha~\shortcite{abacha2019overview}   & Modern Medicine         & General         & -                        & 405;203;383            & NLI;RQE;QA         & Yes                    & English            \\
Tian~\shortcite{tian-etal-2019-chimed}   & Modern Medicine         & General         & -                        & 46,731                 & MRC                & Yes                    & Chinese            \\ \bottomrule
\end{tabular}%
}
\caption{Comparison of medical datasets in traditional and modern medicine. This table only includes textual data. The abbreviations in the table are defined as follows: classification (Class.), machine reading comprehension (MRC), de-identification (De-ID.), disease name recognition (DNR), natural language inference (NLI), recognizing question entailment (RQE), and question answering (QA).}
\label{tab:compare}
\end{table*}
In recent years, health record systems in hospitals have been moving towards digitalization and electronization, and a large amount of clinical data has been accumulated. To make more effective use of these data and provide better medical services, some studies led by MIMIC-III~\cite{johnson2016mimic} have shared these valuable data with medical researchers around the world~\cite{stubbs2015automated,dougan2014ncbi}. Subsequently, with the development of AI, the domain characteristics of various studies have been combined to design various task-oriented datasets~\cite{pampari-etal-2018-emrqa,li-etal-2020-towards,tian-etal-2019-chimed}. These datasets have greatly promoted the development of AI in the medical field and have had a profound impact on society in terms of health and well-being.

\par
However, as shown in Table~\ref{tab:compare}, most of these publicly available datasets focus on modern medicine, there are far fewer datasets on traditional medicine. This is because, compared with traditional medicine, modern medicine has a rigorous, scientific, and standardized medical system, which can efficiently collect high-quality data. Furthermore, the standardization of traditional medicine is still in the development stage, which makes the collection and construction of relevant datasets extremely challenging. Thus the scarce TCM SD datasets has hindered the development of AI in this field. To alleviate this issue, we constructed the first large-scale, publicly available dataset for TCM SD.

\subsection{Natural Language Processing (NLP) in Syndrome Differentiation}
At present, most existing studies have treated SD as a multi-class classification task (i.e., taking the medical records as the input and output the predicted one from numerous candidate syndrome labels). Zhang~\shortcite{zhang2019study} used support vector machines to classify three types of syndromes for stroke patients. Zhang~\shortcite{zhang2020artificial} also introduced an ensemble model consisting of four methods, a back-propagation neural network, the random forest algorithm, a support vector classifier, and the extreme gradient boosting method, to classify common diseases and syndromes simultaneously. Wang~\shortcite{wang2018cnn} proposed a multi-instance, multi-task convolutional neural network (CNN) framework to classify 12 types of syndromes in 1,915 samples. Pang~\shortcite{pang2020effective} proposed a multilayer perceptron (MLP) model with an attention mechanism to predict the syndrome types of acquired immunodeficiency syndrome (AIDS). Similarly, Liu~\shortcite{liu2020end} proposed a text-hierarchical attention network for 1,296 clinical records with 12 kinds of syndromes. However, these approaches only worked well for small-scale datasets. Our work established a series of strong baseline models and conducted comparisons on a larger-scale datasets.
\begin{table}[tb!]
\small
\resizebox{\linewidth}{!}{%
\begin{tabular}{@{}lcccc@{}}
\toprule
\textbf{Model} & \textbf{Corpus} & \textbf{Domain} & \textbf{Language} & \textbf{Corpus Size} \\ \midrule
BERT~\cite{devlin2018bert}          & Wiki+Books      & General         & EN                & 3.3B tokens          \\
RoBERTa-wwm~\cite{cui2021pre}    & Web Crawl       & General         & CN                & 5.4B tokens          \\
MacBERT~\cite{cui-etal-2020-revisiting}        & Web Crawl       & General         & CN                & 5.4B tokens          \\ \midrule
SciBERT~\cite{beltagy2019scibert}        & Web Crawl       & Science         & EN                & 3.2B tokens          \\
BioBERT~\cite{lee2020biobert}        & PubMed          & Medical         & EN                & 4.5B tokens          \\
ClinicalBERT~\cite{alsentzer-etal-2019-publicly}   & MIMIC           & Medical         & EN                & 0.5B tokens          \\
BlueBERT~\cite{peng-etal-2019-transfer}       & PubMed+MIMIC    & Medical         & EN                & 4.5B tokens          \\
PubMedBERT~\cite{gu2021domain}     & PubMed          & Medical         & EN                & 3.1B tokens          \\ \midrule
TCM-BERT$^\star$~\cite{yao2019traditional}        & Web Crawl       & Medical (TCM)   & CN                & 0.02B tokens  \\
ZY-BERT (Ours)           & Web Crawl       & Medical (TCM)   & CN                & 0.4B tokens          \\ \bottomrule
\end{tabular}%
}
\caption{Summary of pre-training details for the various BERT models.}

\label{tab:domain LM}
\end{table}

\subsection{Domain Specific Pre-trained Language Model}
Large-scale neural language models pre-trained on unlabelled text has proved to be a successful approach for various downstream NLP tasks. A representative example is Bidirectional Encoder Representations from Transformers (BERT)~\cite{devlin2018bert}, which has become a foundation block for building task-specific NLP models. However, most works typically focus on pre-training in the general domain, while domain-specific pre-training has not received much attention. Table~\ref{tab:domain LM} summarizes common language models pre-trained in either general domain or specific domain. In general, biomedical and science are mainstream fields of pre-training language model, but in the filed of TCM, there is no much work that has been conducted as for as we know. 
\par
The reasons may be two-fold. On the one hand, TCM lacks large-scale public text corpus, like Wikipedia and PubMed. We deal with this issue by presenting a corpus in TCM domain via crawling and collecting related documents from the websites and books. On the other hand, there is also a lack of downstream tasks that can verify the performance of the pre-training language model, thus we propose the syndrome differentiation task to measure its effectiveness.
\par
To be noticed, an existing work already proposed a language model in the filed of TCM, named as TCM-BERT~\cite{yao2019traditional}, but it did not undergo pre-training of large-scale corpus, but was only finetuned on small-scale nonpublic corpus (0.02B tokens). While, our work provide a more completed TCM-domain corpus (over 20 times larger) and verify its effectiveness during pre-training stage.

\section{Benchmark and Methods}
The TCM-SD benchmark that we collected contains over 65,000 real-world Chinese clinical notes. Table~\ref{tab:example} presents an example. Specifically, each clinical note contains the following five components:
\textbf{Medical history} is the critical information for completing SD. It mainly describes a patient's condition at admission; \textbf{Chief complaint} is a concise statement describing the main symptoms that appeared in the medical history; \textbf{Four diagnostic methods record (FDMR)} is a template statement consisting of four main TCM diagnostic methods: inspection, auscultation and olfaction, interrogation, and palpation; \textbf{ICD-10 index number and name} represents the name and corresponding unique ID of the patient’s disease; \textbf{Syndrome name} is the syndrome of the current patient. However, the raw data could not be used directly for the SD task due to the lack of quality control. Therefore, a careful normalization was further conducted to preprocess the data.

\begin{table*}[t!]
\small
\centering
{\begin{tabular}[c]{|p{0.95\linewidth}|}
\hline
\textbf{Medical History}
\\
The patient began to suffer from  \textcolor{blue}{repeated dizziness} \textcolor{orange}{more than eight years} ago, and the blood pressure measured in a resting-state was higher than normal many times. The highest blood pressure was 180/100 mmHg, and the patient was clearly diagnosed with hypertension. The patient usually took Nifedipine Sustained Release Tablets (20 mg), and the blood pressure was generally controlled, and dizziness occasionally occurred. Four days before the admission, the patient's dizziness worsened after catching a cold, accompanied by asthma, which worsened with activity. Furthermore, the patient coughed yellow and thick sputum. The symptoms were not significantly relieved after taking antihypertensive drugs and antibiotics, and the blood pressure fluctuated wildly. On admission, the patient still experienced dizziness, coughing with yellow mucous phlegm, chills, no fever, no conscious activity disorder, no palpitations, no chest tightness, no chest pain, no sweating, a weak waist and knees, less sleep and more dreams, forgetfulness, dry eyes, vision loss, red hectic cheeks, and dry pharynx, five upset hot, no nausea and vomiting, general eating and sleeping, and normal defecation.
\\
患者\textcolor{orange}{8年余前}开始反复出现 \textcolor{blue}{头晕}，多次于静息状态下测血压高于正常，最高血压180/100 mmHg，明确诊断为高血压，平素服用硝苯地平缓释片20 mg，血压控制一般，头晕时有发作。 此次入院前4天受凉后头晕再发加重，伴憋喘，动则加剧，咳嗽、咳黄浓痰，自服降压药、抗生素症状缓解不明显，血压波动大。 入院时：仍有头晕，咳嗽、咳黄粘痰，畏寒，无发热，无意识活动障碍，无心慌、胸闷，无胸痛、汗出，腰酸膝软，少寐多梦，健忘，两目干涩，视力减退，颧红咽干，五心烦热，无恶心呕吐，饮食睡眠一般，二便正常。  
\\ 
\textbf{Chief Complaint}\\
 \textcolor{blue}{Repeated dizziness} for \textcolor{orange}{more than eight years}, aggravated with asthma for four days.
\\
\textcolor{blue}{反复头晕}\textcolor{orange}{8年余}，加重伴喘憋4天。  
\\ 
\textbf{Four Diagnostic Methods Record}\\ 
Mind: clear; spirit: weak; body shape: moderate; speech: clear,..., tongue: red with little coating; pulse: small and wiry.
\\
神志清晰，精神欠佳，形体适中，语言清晰, ... , 舌红少苔，脉弦细。   
\\
\textbf{ICD-10 Name and ID}: Vertigo (眩晕病) BNG070
\\ 
\textbf{Syndrome Name:} Syndrome of Yin deficiency and Yang hyperactivity 阴虚阳亢证
\\ 
\hline
\textbf{External Knowledge Corpus:}
\\
A syndrome with Yin deficiency and Yang hyperactivity is a type of TCM syndrome. It refers to Yin liquid deficiency and Yang loss restriction and hyperactivity. Common symptoms include  \textcolor{blue}{dizziness}, hot flashes, night sweats, tinnitus, irritability, insomnia, red tongue, less saliva, and wiry pulse. It is mainly caused by old age, exposure to exogenous heat for a long period, the presence of a serious disease \textcolor{orange}{for a long period}, emotional disorders, and unrestrained sexual behavior. Common diseases include insomnia, vertigo, headache, stroke, deafness, tinnitus, premature ejaculation, and other diseases.
\\
阴虚阳亢证，中医病证名。是指阴液亏虚，阳失制约而偏亢，以\textcolor{blue}{头晕目眩}，潮热盗汗，头晕耳鸣，烦躁失眠，舌红少津，脉细数为常见证的证候，多因年老体衰，外感热邪日久，或大病久病\textcolor{orange}{迁延日久}，情志失调，房事不节等所致。常见于不寐、眩晕、头痛、中风、耳聋耳鸣、早泄等疾病中。
\\
\hline
\end{tabular}}
\caption{A sample clinical record from the TCM-SD dataset with related external knowledge. An explicit match between the medical history and external knowledge is marked in blue, while the text in orange is an example of an implicit match that required temporal reasoning.}
\label{tab:example}
\end{table*}

\subsection{Syndrome Normalization}
Like ICD, TCM already has national standards for the classification of TCM diseases, named \textit{Classification and Codes of Diseases and Zheng of Traditional Chinese Medicine} (GB/T15657-1995), which stipulates the coding methods of diseases and the zheng of TCM. However, TCM standardization is still in its early phase of development and faces inadequate publicizing and implementation~\cite{wang2016current}. Some TCM practitioners still have low awareness and different attitudes toward TCM standardization, resulting in inconsistent naming methods for the same syndrome. 

Therefore, based on the above issues, we accomplish syndrome normalization in two stages: merging and pruning.
\paragraph*{Merging} operation is mainly used in two cases. The first is cases in which the current syndrome has multiple names, and all appear in the dataset. For example, \textit{syndrome of wind and heat} (风热证) and \textit{syndrome of wind and heat attacking the external} (风热外袭证) belong to the same syndrome, and we would merge them into one unified name. In this case, we used the national standards for screening. Another is that the current syndrome name does not exist in a standardized form. Therefore, we recruited experts to conduct syndrome differentiation according to the specific case clinical records and finally merge the invalid syndromes into standard syndromes. For example, \textit{syndrome of spleen and kidney yang failure} (脾肾阳衰证) would be merged into \textit{syndrome of spleen and kidney yang deficiency } (脾肾阳虚证).

\paragraph*{Pruning} operation is mainly applied to syndromes with non-standard names that experts fail to differentiate due to vague features. In addition, since syndrome names are hierarchically graded, we pruned out syndromes with higher grades to ensure that the syndromes that appear in the current dataset are the most basic grade, that is the most specific ones that determine the subsequent treatment. For example, \textit{syndrome of wind and cold} (风寒证) is a high-grade syndrome, and its clinical manifestations can be a \textit{syndrome of exterior tightened by wind-cold} (风寒束表证) or \textit{syndrome of wind-cold attacking lung} (风寒袭肺证); each has different symptoms and treatment methods.

\begin{figure*}[ht]
    \includegraphics[width=\linewidth, keepaspectratio]{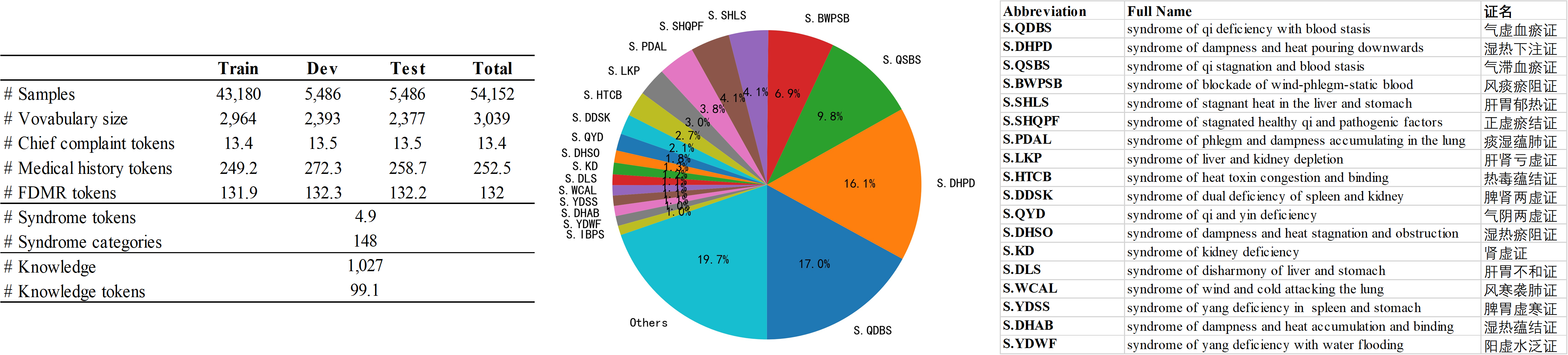}    \caption{The characteristics and syndrome distribution in the dataset.}
    \label{fig:SD}
\end{figure*}
\subsection{Dataset Statistics}
After normalization, the number of syndromes in the dataset was reduced from the original 548 categories to 244.  Considering that some syndromes are infrequent, we further filtered out syndrome categories containing fewer than 10 samples when partitioning the dataset. Then, the processed dataset with 148 syndrome categories and 54,152 samples was divided into a training set, a development (Dev) set, and a test set with a ratio of 8:1:1. The dataset characteristics and syndrome distribution  shown in Figure~\ref{fig:SD}.

\par
Since the data were collected from real-world scenarios, the distribution of syndromes was inevitably unbalanced, leading to a significant gap between the number of rare syndromes and the number of the common ones. The subsequent experiments demonstrate the challenges brought by long-tail distribution issues, and we show that this issue can be mitigated by introducing external knowledge and domain-specific pre-training.

\subsection{External Knowledge}
Current clinical records do not contain any relevant knowledge about the target syndromes, which causes models to have to rely on remembering patterns to complete the task. Therefore, we constructed an external unstructured knowledge corpus encompassing 1,027 types of TCM syndromes by web crawling for information on all the TCM syndromes on the online\footnote{\url{www.dayi.org.cn}}. Specifically, the knowledge of each syndrome consisted of three parts: the cause of the syndrome, the main manifestations, and common related diseases. Table~\ref{tab:example} shows an example. We demonstrate the effectiveness of this knowledge in the experimental section.
\subsection{ZY-BERT}
In general, ZY-BERT differs with TCM-BERT in two main parts: data and pre-training task.

First, the scale and quality of unlabelled text corpus directly affect the performance of pre-trained language models. Previous work TCM-BERT~\cite{yao2019traditional} directly used clinical records as pre-training corpus, resulting in monotonic data type and limited corpus size, which could not meet the needs of large-scale pre-training language model. To deal with this issue, we collected unlabelled data varies in different types from the TCM related websites, including books, articles from websites and academic papers from China National Knowledge Infrastructure (CNKI), counting over 400 million tokens.
\par
Furthermore, the previous work TCM-BERT adopts char masking (CM) and next sentence prediction (NSP) as the pre-training tasks. However, Chinese words usually consist of multiple characters and masking single character might destroy the meaning of the whole word. For example, the word phrase \textit{Yang Deficiency(阳虚)} consists of two characters. Thus, we borrowed the idea of Whole Word Masking from Cui~\shortcite{cui2021pre} and replace NSP with it , which could add challenges to the model training process and allow the model to learn more complex linguistic features. 
\par
Finally, the pre-trained language model consists of 24 Transformer layers, with input dimensionality of 1024. Each transformer contains 16 attention heads. Then we trained the model 300K steps with a maximum learning rate 5e-5 and a batch size of 256. Other hyperparameters and pre-training details are kept same as the ones used in Liu~\shortcite{liu2019roberta}.

\section{Experiments}
We selected the multi-class classification task as the primary form of SD to directly compare the performances of the existing models against the TCM-SD dataset, and used the accuracy and Macro-F1 as evaluation metrics. Specifically, the chief complaint and medical history were concatenated as the inputs, i.e. \textit{[CLS] Chief Complaint [SEP] Medical History [SEP]}, where [CLS] and [SEP] are special tokens used for classification and separation. Then the model predicts the target syndromes from 148 candidate labels based on the representation of [CLS] token. 

\subsection{Baseline}
The baseline methods we used consisted of four types: statistical methods, classical neural-network-based (NN-based) methods, language-model-based (LM-based) methods and domain-specific LM-based methods.
\paragraph{Statistical methods.} These methods were the decision tree (DT) and support vector machine (SVM) methods. These two statistical methods have been widely used in previous studies on SD.
\paragraph{Classical NN-based methods.} These methods included a Bi-LSTM~\cite{schuster1997bidirectional}, a Bi-GRU ~\cite{qing2019novel}, and a two-layer CNN~\cite{kim2014convolutional}. Word embeddings were retrieved from the Chinese version of BERT~\cite{cui2021pre}.
\paragraph{LM-based methods.} These methods included several popular LMs, such as BERT~\cite{devlin2018bert}, RoBERTa~\cite{liu2019roberta}, distillBERT~\cite{sanh2019distilbert}, and ALBERT~\cite{lan2019albert}. These models concatenate multiple pieces of text with special tokens as inputs, make classifications based on the hidden states of the first token, or determine the start and end of the answer by training two classifiers.
\paragraph{Domain-specific LM-based methods.} These methods are similar with LM-based ones but usually pre-trained on domain-specific corpus rather than general domain corpus. TCM-BERT~\cite{yao2019traditional} and our proposed ZY-BERT are the two LM used in this manuscripts.
\begin{table*}[t]
\small

\resizebox{\linewidth}{!}{%
\begin{tabular}{@{}lcccccccc@{}}
\toprule
                & \multicolumn{4}{c}{\textbf{Dev}}                                           & \multicolumn{4}{c}{\textbf{Test}}                                          \\
\textbf{Method} & \textbf{Acc.}    & \textbf{Macro-F1} & \textbf{Macro-R} & \textbf{Macro-P} & \textbf{Acc.}    & \textbf{Macro-F1} & \textbf{Macro-R} & \textbf{Macro-P} \\ \midrule
DT    & 59.42\%       & 20.68\%           & 21.33\%          & 21.52\%          & 59.10\%       & 21.67\%           & 22.38\%          & 22.20\%      \\
SVM           & 77.63\%       & 32.13\%           & 29.56\%          & 43.10\%          & 78.53\%       & 36.37\%           & 32.98\%          & 49.35\%          \\ \midrule
BiLSTM          & 69.30\%          & 17.53\%           & 15.08\%          & 14.76\%          & 69.65\%          & 15.15\%           & 15.65\%          & 17.08\%          \\
BiGRU           & 73.57\%          & 19.53\%           & 20.12\%          & 21.81\%          & 74.43\%          & 20.93\%           & 21.90\%          & 23.76\%          \\
CNN             & 77.56\%          & 31.79\%           & 30.39\%          & 37.99\%          & 78.58\%          & 32.83\%           & 31.29\%          & 39.19\%          \\ \midrule
BERT            & 79.44\%          & 34.18\%           & 34.12\%          & 38.00\%          & 80.17\%          & 35.45\%           & 34.99\%          & 42.00\%          \\
distilBERT      & 79.09\%          & 36.07\%           & 36.62\%          & 38.13\%          & 80.46\%          & 40.24\%           & 39.99\%          & 45.84\%          \\
ALBERT          & 79.62\%          & 37.88\%           & 37.65\%          & 41.94\%          & 80.51\%          & 40.50\%           & 39.57\%          & 46.54\%          \\
RoBERTa          & 80.81\%          & 43.18\%           & 42.55\%          & 47.68\%          &\,\textbf{82.26\%}         & 47.55\%           & 45.72\%          & 54.15\%          \\ \midrule
TCM-BERT          & 79.48\%          & 37.84\%           & 37.60\%          & 42.00\%          & 80.55\%          & 41.58\%           & 40.91\%          & 48.47\%          \\
ZY-BERT(ours)      &\;\;\textbf{81.43\%}$^\dagger$       &\;\;\textbf{49.47\%}$^\dagger$   &\;\;\textbf{48.89\%}$^\dagger$          &\;\;\textbf{54.08\%}$^\dagger$          &\;\,82.19\%$^\dagger$         &\;\;\textbf{51.01\%}$^\dagger$   &\;\;\textbf{49.42\%}$^\dagger$           &\;\;\,\textbf{57.70\% }\!$^\dagger$ \\ \bottomrule
\end{tabular}%
}
\caption{Performance for the classification task. The marker $^\dagger$ refers to $p$-value \textless 0.01. }
\label{tab:classificaiton}
\end{table*}
\subsection{Main Results}
Table~\ref{tab:classificaiton} presents the performances of all the methods for the classification task. Generally, all the methods had good accuracy, which demonstrated that the models were effective at fitting when enough examples were supplied. However, each syndrome in the TCM-SD dataset should have the same importance. Thus, the Macro-F1 is a more accurate metric to evaluate the performances of the models. The Macro-F1 scores achieved by the models were much lower than the accuracy, which demonstrated the challenges of the imbalanced TMC-SD datasets.
\par
Moreover, the statistical methods achieved better scores than the classical NN-based methods. This is because the structures designed for focusing on contextualized representations, such as the Bi-LSTM and Bi-GRU networks, were not good at capturing features, and the performances were worse. In contrast, the SVM and CNN methods were good at extracting local features and obtained better scores. Nonetheless, the language models still achieved the highest scores, demonstrating the effectiveness of the large-scale corpus pre-training.
\section{Discussion}
\subsection{Effect of Domain-specific Pre-training}
The last two rows in Table~\ref{tab:classificaiton} indicates the effects of domain-specific pre-training. To be noticed, our proposed ZY-BERT achieved the astonishing performance improvement and mitigated long-tail distribution issue greatly. On the one hand, Macro-F1 score achieved by ZY-BERT is over 4\% larger than that achieved by RoBERTa, demonstrating the effectiveness of large-scale domain-specific corpus for domain-specific tasks. On the other hand, ZY-BERT also achieves over 10\% Macro-F1 scores higher than the previous domain-specific model TCM-BERT, which proves the quality and reliability of the TCM domain corpus constructed by us.

\subsection{Effect of Knowledge}
\begin{table}[t]
\resizebox{\linewidth}{!}{%
\begin{tabular}{@{}lcccccccc@{}}
\toprule
                & \multicolumn{4}{c}{\textbf{Dev}}                                        & \multicolumn{4}{c}{\textbf{Test}}                                       \\
\textbf{Method} & \textbf{EM} & \textbf{Macro-F1} & \textbf{Macro-R} & \textbf{Macro-P} & \textbf{EM} & \textbf{Macro-F1} & \textbf{Macro-R} & \textbf{Macro-P} \\ \midrule
  \multicolumn{9}{c}{\textbf{Medical History + All Syndromes }}    \\
BERT            & 77.27\%       & 40.71\%           & 41.10\%          & 43.26\%          & 78.20\%       & 45.60\%           & 45.32\%          & 50.15\%          \\
RoBERTa          & 78.71\%       & 45.09\%           & 44.30\%          & 49.38\%          & 80.42\%       & 47.57\%           & 46.42\%          & 51.89\%          \\
  \multicolumn{9}{c}{\textbf{Medical History + Five Syndromes }}    \\
BERT            & 95.59\%       & 77.12\%           & 76.32\%          & 81.04\%          & 95.83\%       & 82.33\%           & 81.35\%          & 86.34\%          \\
RoBERTa          & 95.75\%       & 79.16\%           & 78.74\%          & 82.79\%          & 95.86\%       & 84.42\%           & 84.92\%          & 86.74\%          \\
\multicolumn{9}{c}{\textbf{Medical History + Five Syndromes + Knowledge }}    \\
BERT            & 95.24\%       & 81.21\%  & 81.33\%          & 84.61\%          & 96.06\%       & 85.15\%  & 84.48\%          & 87.92\%          \\
RoBERTa          & \textbf{95.33\%}       &\;\textbf{81.53\%}  &\;\textbf{81.76\%}          &\;\textbf{84.49\% }         & \;\textbf{96.26\%}  &\,\textbf{85.88\%}  &\,\textbf{ 85.59\%  }        & \,\textbf{89.09\%}          \\ \bottomrule
\end{tabular}%
}
\caption{Performance with the machine reading comprehension (MRC) task.}
\label{tab:MRC}
\end{table}
To testify the effectiveness of  the external knowledge corpus, we leveraged knowledge into the model by concatenating the relevant syndrome knowledge with the medical history. However, due to the length limits of the language models, feeding knowledge of all syndromes into the model is infeasible under classification setting. Thus we converted the task from classification to extractive MRC, and designed the following three settings shown in Table~\ref{tab:MRC} to evaluate the significance of the knowledge.

Firstly, we concatenated the original inputs with all syndrome names, and asked the model to extract the target syndrome spans from the context. The competitive results shown between MRC and classification tasks demonstrated that the model had a consistent ability among different task formats without external knowledge. Then we further conducted two groups of experiments. In the first group, instead of concatenating all syndrome names, we only included five syndromes, where one was the target syndrome and the other four were randomly selected. In the second group, we appended the corresponding knowledge for each syndrome selected in the first group. The superior results achieved by the latter group demonstrate the importance of knowledge.
\par
However, the outstanding performance, either with knowledge or without knowledge, was mainly due to the fact that we manually narrowed down the search range to five syndromes. We used the term frequency--inverse document frequency (TFIDF) to search for relevant knowledge from the knowledge corpus based on medical history, and P@5 was only 3.94\%. Thus, knowledge is essential, but finding it is difficult.
\begin{table}[t]

\resizebox{\linewidth}{!}{%
\begin{tabular}{@{}lcccccccc@{}}
\toprule
                & \multicolumn{4}{c}{\textbf{Dev}}                                        & \multicolumn{4}{c}{\textbf{Test}}                                       \\
\textbf{Method} & \textbf{Acc.} & \textbf{Macro-F1} & \textbf{Macro-R} & \textbf{Macro-P} & \textbf{Acc.} & \textbf{Macro-F1} & \textbf{Macro-R} & \textbf{Macro-P} \\ \midrule
  \multicolumn{9}{c}{\textbf{Only Chief Complaint}}    \\
BERT            & 70.56\%       & 23.15\%           & 26.34\%          & 26.34\%          & 71.58\%       & 24.08\%           & 25.38\%          & 24.08\%          \\
RoBERTa          & 71.36\%       & 28.55\%           & 28.85\%          & 33.13\%          & 72.91\%       & 30.78\%           & 34.54\%          & 34.54\%          \\
  \multicolumn{9}{c}{\textbf{Only Medical History }}    \\
BERT            & 79.40\%       & 33.50\%           & 33.46\%          & 37.90\%          & 79.62\%       & 35.57\%      & 35.13\%          & 42.18\%          \\
RoBERTa          & 79.80\%       & 41.40\%           & 40.12\%          & 45.38\%          & 81.83\%       & 45.19\%           & 43.03\%          & 53.78\%          \\
  \multicolumn{9}{c}{\textbf{Chief Complaint + Medical History }}    \\
BERT            & 79.44\%       & 34.18\%  & 34.12\%          & 38.00\%          & 80.17\%       &35.45\%          & 34.99\%          & 42.00\%          \\

RoBERTa          &\;\,\textbf{80.81\%  }     &\,\textbf{43.18\%}  &\;\textbf{42.55\% }         &\;\,\textbf{47.68\% }         &\;\,\textbf{82.26\% }      &\;\,\textbf{47.55\% }        &\;\,\textbf{45.72\% }         &\;\,\textbf{54.15\% }         \\ \bottomrule
\end{tabular}%
}
\caption{Ablation study on the TCM-SD dataset.}
\label{tab:ablation}
\end{table}

\subsection{Ablation Study}
Table~\ref{tab:ablation} shows the results of the ablation study on the TCD-SD dataset. Removing either the medical history or the chief complaint resulted in lower performances, especially if only the chief complaint was taken into account. This was because the chief complaint was typically too short to include sufficient features for classification. However, the chief complaint and medical history complemented each other in a coarse-to-fine fashion.

\subsection{Error Analysis}
By analyzing the error cases, we found that the vast majority of errors occurred in the category with few samples, and fitting only according to the data distribution was still the most significant issue. Except for algorithmic problems, we concluded that there were three main error types:

\paragraph{Complex Reasoning.} As shown in Table~\ref{tab:example}, besides the explicit match marked in blue, there was an implicit match marked in orange that required temporal reasoning. Additionally, the task also included complex reasoning, such as numerical reasoning, spatial reasoning and negative reasoning.
\paragraph{Incomplete Knowledge.} The current models do not take into account the concepts that arise from the SD task, such as Yin and Yang. Therefore, the models do not know how to map the symptoms into the special coordinate system of the TCM diagnostics system.
\paragraph{Out-Of-Vocabulary.} In the clinical records, there exists not only academic medical-related terms but also various rare traditional characters in TCM, which impeded the understanding of the context.

\section{Conclusions}
This paper introduced a meaningful task, SD, in TCM and its connection with NLP and presented the first public large-scale benchmark of SD: TCM-SD. Furthermore, a knowledge corpus supporting the model understanding and the large-scale TCM domain corpus for pre-training were constructed. Moreover, one domain-specific pre-training language model named as ZY-BERT was proposed. The experiments on this dataset demonstrated the challenges, the inadequacy of existing models, the importance of knowledge and the effectiveness of domain-specific pre-training. This work can greatly promote the internationalization and modernization of TCM, the proposed benchmark and associated baseline models  provide a basis for subsequent research.

\end{CJK*}
\section*{Acknowledgements}
This work is supported by funds from the National Natural Science Foundation of China (No.U21B2009). The data used in this paper were only routine diagnosis and treatment data of patients, excluding any personal information of the patients (such as name, age, and telephone number). This study did not interfere with normal medical procedures or create an additional burden to medical staff, and no experiments were conducted on patients. \textbf{All the data have been desensitized}. Therefore, this paper does not involve ethical issues and waives the requirement of individual patient consent. We public TCM-SD dataset, TCM-domain corpus and ZY-BERT model at \url{https://github.com/Borororo/ZY-BERT}. We thank the reviewers for their helpful and constructive comments. And we thank M.D. Yonglan Zhou for her insightful and professional suggestions.


\bibliographystyle{ccl}
\bibliography{ccl2022-en}

\end{document}